\pdfoutput=1

\documentclass[11pt]{article}

\usepackage{misc/acl}

\usepackage[utf8]{inputenc} 
\usepackage[T1]{fontenc}    
\usepackage{times}          
\usepackage{inconsolata}    
\usepackage{pifont}

\usepackage{microtype}      
\usepackage{url}            
\usepackage{hyperref}       

\usepackage{rotating}
\usepackage{makecell}
\usepackage{booktabs}       
\usepackage{array}          
\newcolumntype{C}[1]{>{\centering\arraybackslash}p{#1}}

\usepackage{multirow}       
\usepackage{adjustbox}      
\usepackage{colortbl}       
\definecolor{Gray}{gray}{0.98}
\newcolumntype{g}{>{\columncolor{Gray}}c}
\usepackage{tabularx}

\usepackage{xcolor}

\usepackage{amsmath}        
\usepackage{amssymb}        
\usepackage{amsfonts}       
\usepackage{mathtools}      
\usepackage{amsthm}         
\usepackage{bm}
\usepackage{esvect}

\usepackage{cleveref}

\usepackage[inline, shortlabels]{enumitem} 

\usepackage{graphicx}       
\usepackage{subfigure}      
\usepackage{xspace}         
\usepackage{svg}
\usepackage{subcaption}

\usepackage{listings}
\usepackage{caption}
\lstdefinestyle{fancyquote}{
    frame=none,
    backgroundcolor=\color{gray!10},
    breaklines=true,
    columns=fullflexible,
    showstringspaces=false,
    keepspaces=true,
    xleftmargin=2em,
    xrightmargin=2em,
    framexleftmargin=1em, 
    framexrightmargin=1em, 
    aboveskip=2em,        
    belowskip=2em,         
    moredelim=**[is][\color{teal}]{@}{@} 
}

\usepackage[algoruled,ruled,vlined,noend]{algorithm2e} 

\usepackage{soul}           
\usepackage{changepage}     
\usepackage{CJK}            

\usepackage{todonotes}

\definecolor{TodoColor}{rgb}{1,0.7,0.6}
\definecolor{TodoColor2}{rgb}{0.7,0.7,0.9}
\definecolor{TodoColor3}{rgb}{0.5,0.8,0.5}

\newcommand{\exht}{ZS-Exp$_{M\leftrightarrow H}$\xspace}
\newcommand{\expt}{ZS-Exp$_{H_\alpha \leftrightarrow H_\beta}$\xspace}
\newcommand{\contht}{SAE Cont.$_{M\leftrightarrow H}$\xspace}
\newcommand{\contpt}{SAE Cont.$_{H_\alpha \leftrightarrow H_\beta}$\xspace}
\newcommand{\pflip}{P~{\footnotesize $_\textsc{flip}$}}

\definecolor{darkgreen}{HTML}{466362}          
\newcommand{\Hone}{\textcolor{darkgreen}{H1}}
\definecolor{lightgreen}{HTML}{84BC9C}          
\newcommand{\Htwo}{\textcolor{lightgreen}{H2}}
\definecolor{pastelorange}{HTML}{CB793A}          
\newcommand{\MT}{\textcolor{pastelorange}{MT}}

\definecolor{novelC}{HTML}{96d6d6}        
\definecolor{promptingC}{HTML}{E06666}
\definecolor{steeringC}{HTML}{9D54CE}

\newcommand{\emojiWord}[1]{\raisebox{-0.2ex}{\includegraphics[height=1em]{#1}}}
\newcommand{\classifier}{\emojiWord{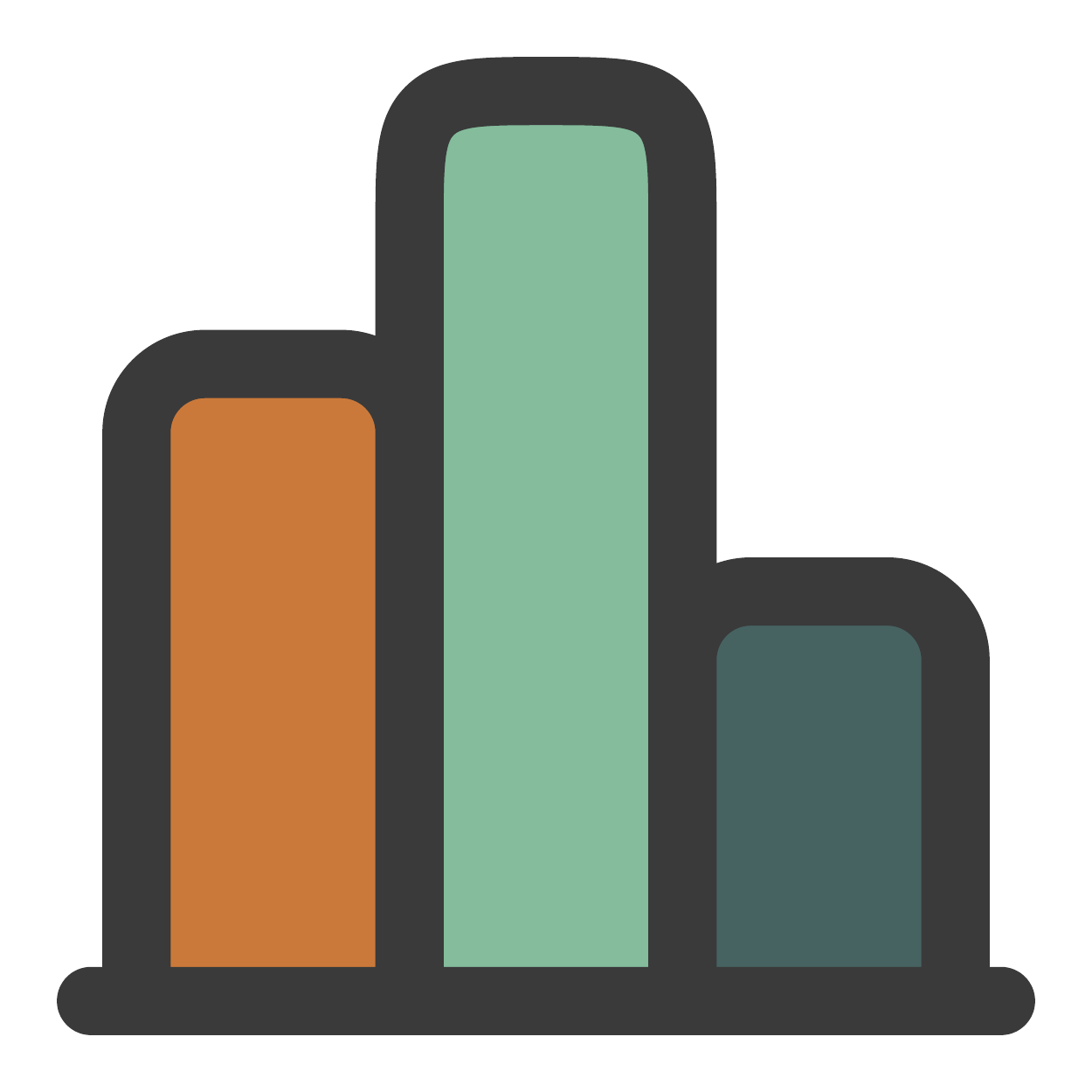}~classifier}
\newcommand{\comet}{\mbox{\emojiWord{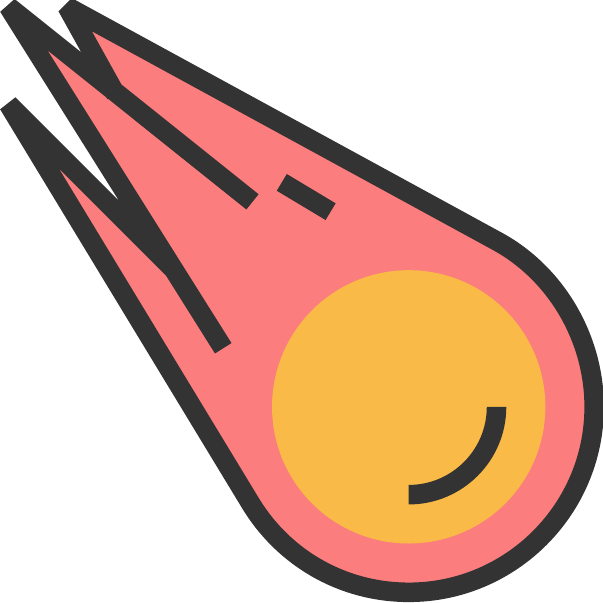}\kern0.1em~Comet}}

\graphicspath{{img/}}

\hypersetup{
}

\title{Steering Large Language Models for Machine Translation Personalization}

\author{
  Daniel Scalena$^{1,2}$\thanks{~~Equal contribution.} ~\,~ Gabriele Sarti$^{1*}$ \\ \textbf{Arianna Bisazza}$^1$ ~\,~ \textbf{Elisabetta Fersini}$^2$ ~\,~  \textbf{Malvina Nissim}$^1$ \vspace{3mm}\\
  $^1$CLCG, University of Groningen ~\,~ $^2$University of Milano-Bicocca \vspace{3mm}\\
  \small{\texttt{d.scalena@campus.unimib.it} ~\,~ \texttt{g.sarti@rug.nl}}
}

\begin{document}
\maketitle
\begin{CJK*}{UTF8}{gbsn}

\begin{abstract}
Large language models have simplified the production of personalized translations reflecting predefined stylistic constraints. However, these systems still struggle when stylistic requirements are implicitly represented by a set of examples, such as texts produced by a specific human translator. In this work, we explore various strategies for personalizing automatically generated translations when few examples are available, with a focus on the challenging domain of literary translation. We begin by determining the feasibility of the task and how style information is encoded within model representations. Then, we evaluate various prompting strategies and inference-time interventions for steering model generations towards a personalized style, with a particular focus on contrastive steering with sparse autoencoder (SAE) latents to identify salient personalization properties. We demonstrate that contrastive SAE steering yields robust style conditioning and translation quality, resulting in higher inference-time computational efficiency than prompting approaches. We further examine the impact of steering on model activations, finding that layers encoding personalization properties are impacted similarly by prompting and SAE steering, suggesting a similar mechanism at play.
\end{abstract}

\section{Introduction}


\begin{figure}[t]
    \centering
    \includegraphics[width=1\linewidth]{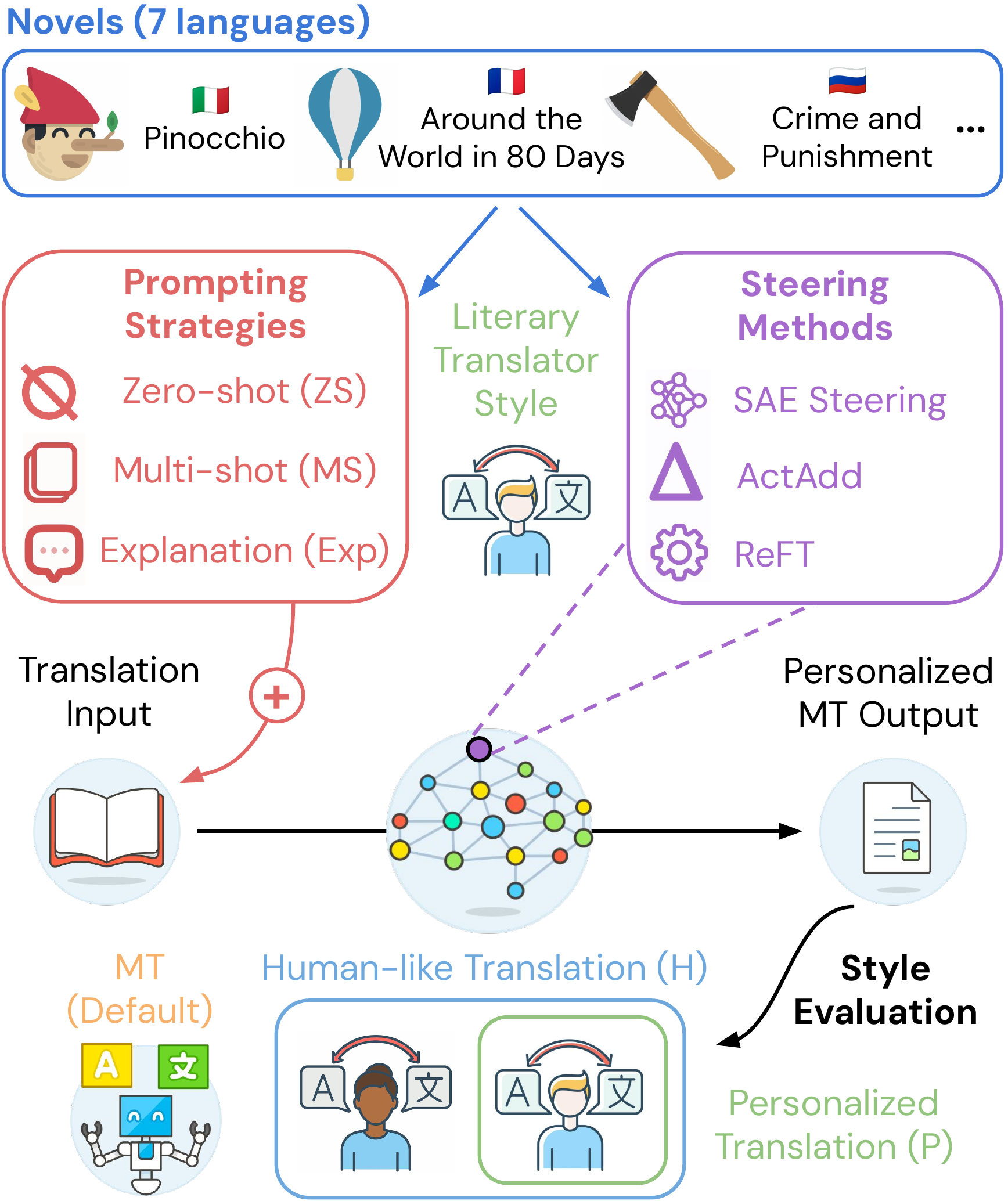}
    \caption{We compare \textcolor{promptingC}{prompt-based approaches} with \textcolor{steeringC}{steering techniques} intervening on model internals for personalizing MT outputs in literary machine translation. We use MT quality metrics and style classifiers to quantify the impact of steering on output fluency and personalization accuracy.}
    \label{fig:teaser}
    \vspace{-12pt}
\end{figure}

A translated novel is not a simple literal retelling of the story---the personal style of the translator and their lexical and stylistic choices play a crucial role in shaping the narrative in a new language. Past efforts in the automatic translation of literary works have historically been constrained by the limited capabilities and flexibility of machine translation (MT) systems. The recent popularization of MT systems based on large language models (LLMs) has dramatically improved their capacity in handling the long contexts typical of literary translations, but mimicking the creative and rich language that characterizes the translators' own style remains an open issue. While prompting and tuning-based strategies have been successfully used for MT personalization~\citep{michel-neubig-2018-extreme,wang-etal-2021-towards}, recent advances in interpretability research have highlighted the effectiveness of various \textit{steering} methods, which employ surgical interventions on LLMs' activations to condition model generation~\citep{rimsky-etal-2024-steering}. While steering techniques were primarily applied to explicit properties such as sentiment and formality, this work aims to test their effectiveness for MT personalization. We compare steering methods with established prompting techniques, with a particular focus on the challenging domain of literary translation, where stylistic choices are more evident.

Provided that stylistic attribution is notoriously challenging for human annotators \citep{youyou2015computer,flekova-etal-2016-analyzing}, we begin with a preliminary assessment to verify whether trained classifiers can distinguish between translators' styles and whether LLMs can use demonstrations to enhance the stylistic accuracy of their translations. We then connect the conditioning induced by prompting to the inner workings of the model, identifying activations with high discriminative capacity for style differences in intermediate layers of LLMs.
In light of our results indicating that style information is effectively detected and represented by automatic systems, we evaluate the effectiveness of various established prompting and steering methods on three multilingual LLMs across novels in seven mid- to high-resourced languages. We focus in particular on a contrastive steering approach using sparse autoencoders (SAEs,~\citealp{cunningham2023sparseautoencodershighlyinterpretable}) to condition model generations by upweighting sparse style-related latents at inference time.

Our results demonstrate that contrastive SAE steering is a promising approach for MT personalization, yielding translations that align more closely with general human translation features and exhibit a greater similarity with the desired personalized style compared to other methods. Importantly, these results are achieved with no degradation in translation quality, assessed according to established MT quality metrics. We conclude by comparing the impact of our method on model representations with the outcome of multi-shot prompting, finding that probes trained on prompt-conditioned activations can predict the effectiveness of SAE steering with high precision. These results confirm that prompting and SAE steering techniques converge to similar solutions for conditioning model behavior, enabling future investigations into the mechanistic impact of prompting through the study of interpretable SAE latents.

\section{Related Work}
\label{sec:sota}

\paragraph{Machine Translation of Literary Texts} The literary domain has historically been challenging for automatic MT systems due to their limited ability in handling rich linguistic and cultural contexts~\citep{matusov-2019-challenges} and their propensity to produce overly literal outputs~\citep{guerberof-toral-2022-creativity}. Automatic literary translation has a long history dating back to pre-neural MT approaches~\citep{voigt-jurafsky-2012-towards,toral-way-2015-translating,Toral2018,moorkens-etal-2018-translators} with two recent dedicated evaluation campaigns~\citep{wang-etal-2023-findings,wang-etal-2024-findings}. The advent of LLMs brought new opportunities in the processing of longer context for document-level translation~\citep{wang-etal-2023-document-level,briakou-etal-2024-translating,wu2025perhapshumantranslationharnessing}, but critical errors requiring human translator's intervention nonetheless persist~\citep{karpinska-iyyer-2023-large}. In this work, we use the \textsc{par3} dataset~\citep{thai-etal-2022-exploring}, which comprises multiple human translations of novels to evaluate MT personalization in the literary domain.

\paragraph{Personalization for Machine Translation} Advances in MT quality recently led to a growing interest in personalization approaches to ensure a consistent format and appropriate stylistic choices in model generations~\citep{rabinovich-etal-2017-personalized,lin-etal-2021-towards}. Previous techniques for controlling attributes such as formality~\citep{sennrich-etal-2016-controlling,niu-etal-2017-study,nadejde-etal-2022-cocoa} or gender~\citep{vanmassenhove-etal-2018-getting,saunders-byrne-2020-reducing} typically required tuning existing models on pre-defined properties of interest, with few works attempting a real data-driven adaptation from unlabeled demonstrations~\citep{michel-neubig-2018-extreme,wang-etal-2021-towards,zhang-etal-2022-building}. More recently, several studies employed prompting~\citep{garcia2022usingnaturallanguageprompts,sarti-etal-2023-ramp} or preference optimization from post-editing behavior~\citep{lee-etal-2023-pepe,berger-etal-2024-post} to render MT personalization more effective and data-efficient. In this work, we complement prompt results with steering approaches to personalize MT outputs using a few user-provided examples.

\paragraph{Steering Language Model Generations} Steering approaches exploit the linear structure of LM activations~\citep{mikolov-etal-2013-linguistic,chanin-etal-2024-identifying} to craft inference-time interventions for influencing model generations. These methods commonly employ contrastive sets of in-context demonstrations~\citep{rimsky-etal-2024-steering,scalena-etal-2024-multi} to map input properties to components such as vectors~\citep{actadd,li-etal-2023-iti}, linear probes~\citep{repe}, or learned projections~\citep{wu2024reft,wu2025axbenchsteeringllmssimple}. Sparse Autoencoders (SAEs) are another family of promising approaches for enabling fine-grained interventions in language models~\citep{yun-etal-2021-transformer,cunningham2023sparseautoencodershighlyinterpretable,templeton2024scaling}. They are trained to decompose activations into approximately monosemantic features, offering a potentially interpretable basis for modifying model behavior. While interpreting their learned latents remains non-trivial~\citep{marks2025sparse}, SAEs have proven effective for applying targeted interventions along specific linear directions~\citep{chalnev2024improvingsteeringvectorstargeting,zhao-etal-2025-steering,ferrando2025iknowentityknowledge}. However, most research on SAEs has so far focused on synthetic tasks or standard benchmarks, leaving their potential in real-world settings relatively underexplored.

\section{Preliminaries}
\label{sec:preliminary}

\begin{table}
\centering
\small
\begin{tabular}{lp{11em}c}
\toprule
\textbf{Lang.} & \textbf{Novel} & \textbf{Train / Val / Test} \\ 
\midrule
IT & Pinocchio                          & 745 / 82 / 107 \\
FR & Around the World in Eighty Days    & 829 / 92 / 120 \\ 
NL & The Diary of a Young Girl          & 769 / 85 / 110 \\ 
DE & Beware of Pity                     & 606 / 67 / 96 \\ 
RU & Crime and Punishment               & 1517 / 168 / 224 \\ 
JA & No Longer Human                    & 652 / 40 / 81 \\ 
ZH & Dream of the Red Chamber           & 694 / 76 / 92 \\ 
\bottomrule
\end{tabular}
\caption{\label{tab:novels-details}Novels and number of example paragraphs used in our evaluation. Each example corresponds to a triple (\MT, \Hone, \Htwo). 20 additional examples per language are reserved for prompting/steering purposes.}
\end{table}

\begin{table}
    \centering
    \small
    \scalebox{0.9}{
\begin{tabular}{p{0.24\linewidth}@{}p{0.83\linewidth}}
\toprule[1.5pt]
\multicolumn{2}{p{0.90\linewidth}}{\textsc{zh}~$\rightarrow$~\textsc{en} ~ 道人道：「既如此，便隨你去來。」} \\
\midrule
\Hone & \textit{``Such being the case,'' the Taoist acquiesced, ``I am ready to follow you, whenever you please to go.''} \\
\Htwo & \textit{``Very good, I will go with you then,'' said the Taoist.} \\
\midrule
\textcolor{pastelorange}{Gemma 2~$_{\text{2B}}$} & \textit{``If that's the case, then go ahead'' said the Taoist.} \\[4pt] 
\textcolor{pastelorange}{Gemma 2~$_{\text{9B}}$} & \textit{``If so, then I will go with you'' the Taoist said} \\[4pt] 
\textcolor{pastelorange}{Llama 3.1~$_{\text{8B}}$} & \textit{The Taoist said:``If you insist on going, go ahead.''} \\[4pt] 
\midrule
\toprule
\multicolumn{2}{p{0.90\linewidth}}{\textsc{it}~$\rightarrow$~\textsc{en} ~ Sar\`a l'effetto dell'acqua del mare. Il mare ne fa di questi scherzi} \\
\midrule
\Hone & \textit{"The salt water must have done it. The sea plays funny tricks."} \\
\Htwo & \textit{"It must have been the effect of sea-water. The sea makes extraordinary changes."} \\
\midrule
\textcolor{pastelorange}{Gemma 2~$_{\text{2B}}$} & \textit{"It will be the effect of the sea water. The sea makes of these jokes."} \\[4pt] 
\textcolor{pastelorange}{Gemma 2~$_{\text{9B}}$} & \textit{It will be the effect of the sea water. The sea plays these tricks.} \\[4pt] 
\textcolor{pastelorange}{Llama 3.1~$_{\text{8B}}$} & \textit{It will be the effect of the sea water. The sea does things like this.} \\[4pt] 

\bottomrule[1.5pt]
\end{tabular}
}
    \caption{Examples from \textsc{par3} for \textsc{zh}~$\rightarrow$~\textsc{en} (\textit{``Dream of the Red Chamber''} by Cao Xueqin) and \textsc{it}~$\rightarrow$~\textsc{en} (\textit{``The Adventures of Pinocchio''} by Carlo Collodi), including two human translations (\Hone, \Htwo) and \textcolor{pastelorange}{LLM outputs} with zero-shot prompting. More examples in~\Cref{app:output-examples}.}
    \label{tab:data-examples}
    \vspace{-12pt}
\end{table}

We begin our investigation by validating some key assumptions: \textbf{i)} Whether the personal translation styles are \textit{discernible}, i.e., if the translation style of various LLMs and human translators can be consistently identified; \textbf{ii)} Whether LLMs can mimic specific translators' styles when provided with some of their translations; and \textbf{iii)} Whether style distinctions are reflected in the model's internal representations, to motivate the interest in steering approaches for improving MT personalization.

In our experiments, we use the \textsc{par3} dataset by~\citet{thai-etal-2022-exploring}, which comprises multiple human translations of novels from seven diverse languages (German, Russian, Chinese, Italian, Dutch, French, and Japanese) into English, as a benchmark for evaluating MT personalization. Novels are segmented into paragraphs with translations into English by two professional literary translators. \Cref{tab:novels-details} shows statistics about the \textsc{par3} data, while \Cref{tab:data-examples} presents some examples for Chinese-to-English and Italian-to-English. We name the two available human translations \Hone~and \Htwo, and compare them with MT outputs produced by LLMs, which we denote as \MT$_{\text{model}}$ from here onwards. We evaluate three LLMs, namely Llama 3.1 8B Instruct~\citep{grattafiori2024llama3herdmodels} and Gemma 2~\citep{gemmateam2024gemma2improvingopen} in its 2B and 9B instruction-tuned variants. Our model selection is motivated by our steering requirements, discussed in~\Cref{sec:exp}.

\subsection{Is MT Personalization Discernible?}
\label{sec:discern}


Following prior work on personalization~\citep{wang-etal-2024-m4gt,liu-etal-2023-coco}, we train a series of \classifier{}s based on multilingual XLM Transformer encoders~\citep{conneau-etal-2020-unsupervised} to distinguish between \Hone, \Htwo, and \MT~translations for each language and each MT model in our evaluation suite. If classifiers can accurately predict the translation style among these three styles, then this provides evidence for distinct stylistic signals between them. In particular, the ability to distinguish between \Hone~and \Htwo~would suggest that style signals can help differentiate between different human translators, rather than simply between human-like and automatic translations.\footnote{Details in~\Cref{app:classifiers}. Full results in~\Cref{app:classifier-results}.} We find that the style information in the provided examples can be readily identified, with all \classifier{}s reaching an accuracy between 77\% for Japanese and 99\% for Chinese, or 86\% on average. These results corroborate previous evidence on the ability of automatic systems in detecting stylistic clues, while the task remains elusive for human annotators~\citep{youyou2015computer,flekova-etal-2016-analyzing,wang-etal-2024-semeval-2024}.



\subsection{Can LLMs Mimic Translation Styles?}
\label{sec:produce}
We test the ability of LLMs to mimic the stylistic choices of a particular translator in a traditional multi-shot (MS) prompting setup. For each translator available across all tested novels, we prompt the model with 20 in-context examples selected from the original pool of translated paragraphs by that translator, asking it to generate new translations matching the provided style. We compare MS results with the default zero-shot (ZS) prompting using our high-scoring \classifier{}s to quantify the effect of in-context examples on personalization accuracy, and using the popular \comet{} MT metric~\citep{rei-etal-2020-comet} to track translation quality.~\Cref{tab:zs-ms-perf} presents our results.  The proportion of outputs categorized as matching the translator's style is increased two- to four-fold when in-context examples are provided, confirming that LLMs can employ implicit clues in small sets of user examples to improve the style accuracy of their translations. Stable scores for \comet{} also indicate that translation quality is preserved after style adaptation. 




\begin{table}
    \small
    \centering
    \begin{tabular}{l|cccccc}
    \toprule[1.5pt]
     & \multicolumn{2}{c}{\textbf{Gemma 2 2B}} & \multicolumn{2}{c}{\textbf{Gemma 2 9B}} & \multicolumn{2}{c}{\textbf{Llama 3.1 8B}} \\
    \cmidrule(lr){2-3}
    \cmidrule(lr){4-5}
    \cmidrule(lr){6-7}
    & \emojiWord{assets/chart-classifier.pdf} & \textbf{\emojiWord{assets/comet3.pdf}} & \emojiWord{assets/chart-classifier.pdf} & \textbf{\emojiWord{assets/comet3.pdf}} & \emojiWord{assets/chart-classifier.pdf} & \textbf{\emojiWord{assets/comet3.pdf}} \\
    \midrule
    ZS & 0.10 & 0.69 & 0.08 & 0.71 & 0.08 & 0.70 \\
    MS & \textbf{0.24} & 0.69 & \textbf{0.31} & 0.73 & \textbf{0.32} & 0.73 \\
    \bottomrule[1.5pt]
    \end{tabular}
    \caption{Classifier-based personalization accuracy (\emojiWord{assets/chart-classifier.pdf}) and Comet-based translation quality (\emojiWord{assets/comet3.pdf}) for zero-shot (ZS) and multi-shot (MS) prompting with 20 in-context examples averaged across all translators and languages.}
    \label{tab:zs-ms-perf}
    \vspace{-4pt}
\end{table}

\subsection{Is Style Detectable in Model Activations?}
\label{sec:probing}

\begin{figure}
    \centering
    \includegraphics[width=.95\linewidth]{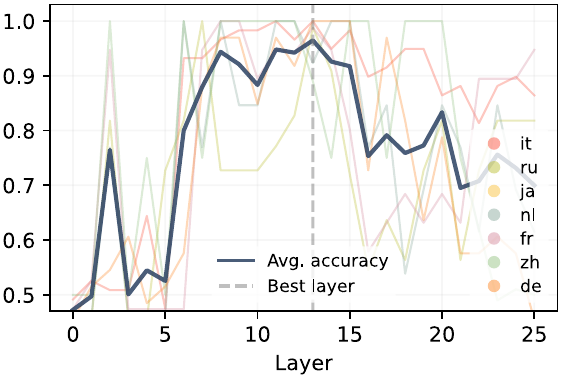}
    \includegraphics[width=1\linewidth]{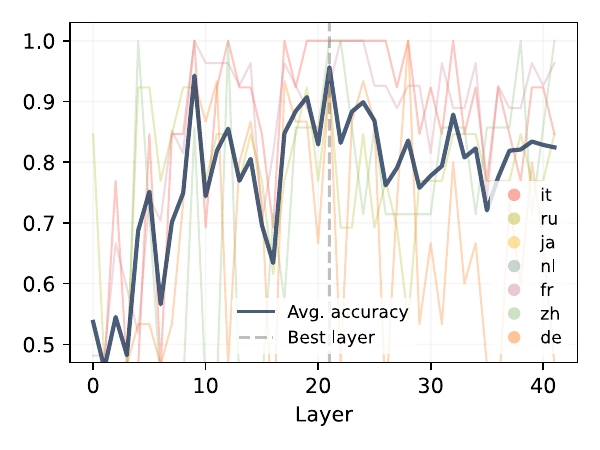}
    \caption{Probing classifier performance on the human translation detection task across Gemma 2 2B (top) and 9B (bottom) layers. Activations in intermediate layers are found to capture translation style information with high precision.
    }
    \label{fig:probing_both}
    \vspace{-12pt}
\end{figure}

In light of these results, we set out to test where the stylistic information enabling the improved performance of MS prompting is encoded within model representations. To this purpose, we train \textit{linear probes}~\citep{belinkov-2022-probing} using the last prompt token activations as input features to predict a binary class (\MT~or Human) and comparing it to the label the \classifier{} (from \Cref{sec:discern}) would assign to the resulting LLM translation.\footnote{To simplify the probing setup, the probe Human class is used for both \Hone~and \Htwo~classifier predictions.} High probing accuracy in this setting would suggest that the style information encoded from prompt examples is sufficient to predict the appropriateness of model translations before they are even generated. The training set of our probes comprises examples with 20 in-context demonstrations, where half of the examples use personalized translations made by a human translator, and the other half uses MT-generated demonstrations we pre-generated using the same model in a ZS setup. For each novel, we select as test examples only left-out entries for which the \classifier{} predictions flip from \MT~in ZS setting to the desired style (\Hone or \Htwo) when demonstrations are provided, to focus our analysis on settings where in-context demonstrations produce a tangible effect on output personalization.\footnote{In-context demonstrations are resampled for every test paragraph to prevent overfitting on spurious features.} This balanced setup precludes shortcut learning from spurious information to trained probes, e.g., the number of prompt demonstrations, ensuring that stylistic differences between human-made and MT-generated demonstrations are the sole distinguishing factor leading to discrepancies in model activations.

We focus specifically on Gemma models, extracting activations after the attention block at each model layer for the last token of the prompt, which was previously shown to encode key task-relevant information~\citep{hendel-etal-2023-context,todd2024function,scalena-etal-2024-multi}.~\Cref{fig:probing_both} reports probe accuracies across all Gemma 2 2B layers, with results for the 9B model reported in~\Cref{app:all-models}. We find a peak in probe accuracy of $\sim$95\% around intermediate model layers, suggesting that these layers encode stylistic information with very high precision.\footnote{We find probes for layers 13 and 21 to perform best for the 2B and 9B models, respectively.} These results confirm that personalization features are clearly discernible from LLMs' activations, motivating our following experiments on activation steering for MT personalization.

\section{Methods}

We now introduce the methods evaluated in our experiments, with a focus on our proposed SAE-based contrastive steering method. In all our contrastive formulations, we treat the available \Hone~and \Htwo~translations as distinct personalization targets to contrast with \MT, and use $H^+, H^-$ to indicate desired and undesired human styles, respectively.\footnote{E.g., when mimicking \Hone~style, $H^+$=~\Hone~and $H^-$=~\Htwo.}

\subsection{Prompting Baselines}
\label{sec:prompting}

\paragraph{Zero-Shot (ZS).} The ZS setup of our main experiment matches the one from~\Cref{sec:discern}, in which the model is asked to produce a translation with no conditioning from examples or explanations towards the target translation style. We use this setting to establish a baseline style and translation quality performance for tested models.

\paragraph{Zero-Shot with Explanation (ZS-Exp).}
We extend the ZS setting by exploiting verbalized style description produced by  GPT-4o~\cite{openai2024gpt4technicalreport}. We prompt GPT-4o with 20 $H^+$ demonstrations and 20 \MT~outputs produced in the ZS settings, asking it to synthesize a detailed explanation of salient elements characterizing the desired style.\footnote{Prompt templates in Listings \ref{lst:explain-template}, \ref{lst:explain-example},\ref{lst:explain-final} of \Cref{app:explain}.} We manually review all generated explanations to ensure they do not contain any verbatim excerpts from input examples, and use them to the tested LLM in a setting with no demonstrations (ZS). This setup allows us to verify whether natural language directives can act as effective and interpretable alternatives to other tested methods.

\paragraph{Multi-Shot (MS).}
Following~\Cref{sec:produce}'s findings, we adopt the same MS setup using 20 in-context translation examples matching the style of a desired human translator $H^+$.


\subsection{Steering Baselines}

\paragraph{Activation Addition (ActAdd).} ActAdd is a simple yet effective technique for steering language models. We employ the standard formulation by~\citet{rimsky-etal-2024-steering}, contrasting a set of \textit{positive} activations $\left\{z\right\}^+$, here the style-aligned $H^+$ translations, with \textit{negative} activations $\left\{z\right\}^-$, here the default \MT~output. We perform steering on the style-relevant layers identified in~\Cref{sec:probing} by computing the average $\Delta$ steering vector as the difference-in-means for the two sets of activations using 20 in-context examples, and applying it additively to the same model layer during inference. We use a factor $\alpha = 2$ to control steering intensity, as this was found effective in previous research~\citep{scalena-etal-2024-multi}.

\paragraph{Representation Fine-tuning (ReFT).} ReFT was recently introduced as an alternative to traditional weight-based Parameter-Efficient Fine-Tuning (PEFT) techniques, aiming to learn task-specific interventions applied directly to model activations at inference time \citep{wu2024reft}. As for ActAdd, we apply ReFT to the style-relevant layers identified in~\Cref{sec:probing} and limit confounding factors by tuning ReFT interventions with the same set of 20 examples used for MS prompting.

\subsection{Contrastive SAE Steering (SAE Cont.)}

We now introduce our proposed contrastive formulation for steering translation style with SAE latents.
Given the set of paragraphs $\mathcal{D}$ for a novel in the \textsc{par3} dataset, each instance in it is a tuple $\left< s, H^+, H^-, \text{\MT} \right>$,  where $s$ is the non-English source sentence. Similar to previous methods, we employ a contrastive approach to extract SAE latents that are most active in the presence of the selected translator's style, while simultaneously controlling for generic features occurring in natural model outputs.

\paragraph{Feature extraction}  First, we gather activations $z^+_l$ and $z^-_l$ by prompting the model with tuples $\left< s, H^+ \right>$, $\left< s, \text{\MT} \right>$, respectively. Activations are extracted at the last prompt token position from the most style-related layer, as identified in Section~\ref{sec:probing}. Activations are then converted into sparse feature dictionaries $x^{+} = h(z^+)$ and $x^{-} = h(z^-)$, with $x^+, x^- \in \mathbb{R}^m$ and $h$ representing the the SAE encoder module. This procedure is repeated separately across 20 contrastive examples, resulting in two collections of SAE latent vectors for positive/negative examples, $\mathcal{X}^+ = \left\{x^{+}_1, x^{+}_2, \dots, x^{+}_{20} \right\}$ and $\mathcal{X}^- = \left\{x^{-}_1, x^{-}_2, \dots, x^{-}_{20} \right\}$

\paragraph{Relevance-based Feature Selection} To identify discriminative features for personalization in the large set of latents, we adapt the information-theoretic approach proposed by~\citet{zhao-etal-2025-steering}. For each of the inputs, we identify the subset of size $n \ll m$ of SAE active features, i.e. latent dimensions for which the logit is $> 0$. We consider logit values in this subset as instances of a random variable $X_i \in x$, and calculate the mutual information $I(X_i, Y)$ between each feature $X_i$ and the target binary variable $Y = \left\{ +, - \right\}$, representing whether the text matches the desired $H^+$ style or not. A higher $I(X_i, Y)$ indicates that the $i$-th feature is more informative for discriminating between personalized and default inputs, and can hence be used for steering. A representative sample of 40 latents showing the highest mutual information scores for both personalized ($\left\{ X_i \right\}^+$) and non-personalized ($\left\{ X_i \right\}^-$) examples is selected using this procedure. This procedure differs from traditional SAE-based steering methods, which employ only features associated with the positive class~\citep{chalnev2024improvingsteeringvectorstargeting, NEURIPS2024_f5454485}, in that it encourages the selection of discriminative features, both positive and negative. For every selected latent, we use its expected logit when personalization is present or absent, given the set of provided examples, i.e. $\mathbb{E}^+[X_i]$ and $\mathbb{E}^-[X_i]$.



\paragraph{Inference-time intervention} Finally, activations are steered by clipping selected latents' values to their expected value, hence promoting $H^+$-aligned latents $\left\{ X_i \right\}^{H^+}$ while demoting \MT-aligned ones $\left\{ X_i \right\}^{\text{\MT}}$.\footnote{Algorithm \ref{alg:latent-steering} summarizes our SAE steering approach.} Additionally, we modulate the magnitude of the resulting vector with an $\alpha$ \textit{coefficient}, which was found  to play an essential role in steering effectiveness in previous research~\citep{scalena-etal-2024-multi,ferrando2025iknowentityknowledge}

%

\section{Experiments}
\label{sec:exp}

\subsection{Setup}

\paragraph{Model selection}
We evaluate our methods on the Llama 3.1 and Gemma models previously used in~\Cref{sec:preliminary}. Our selection is guided by the availability of open-source pre-trained SAEs, which can be otherwise computationally expensive to train. For Gemma models, we employ SAEs from the GemmaScope suite~\citep{lieberum-etal-2024-gemma}, while for the Llama~3.1 model we use the SAE by~\citet{goodfire_sae}. GemmaScope SAEs are available for every model layer, enabling us to steer Gemma models on their most informative layers for the task, which we identified in \Cref{sec:probing}. On the contrary, a single SAE for the 19th layer is available for Llama, hence limiting our evaluation of SAE steering and potentially producing sub-optimal steering results for that model.

\begin{table*}[t]
    \small
    \centering
    \resizebox{\textwidth}{!}{%
    \begin{tabular}{l ccccc|ccccc|ccccc}
    \toprule[1.5pt]
        & \multicolumn{5}{c}{\textbf{Gemma 2 2B}} & \multicolumn{5}{c}{\textbf{Gemma 2 9B}} & \multicolumn{5}{c}{\textbf{LLaMA 3.1 8B}} \\ 
        \cmidrule(lr){2-6}
        \cmidrule(lr){7-11}
        \cmidrule(lr){12-16 }
        & \textbf{H} & \textbf{P} & \textbf{\pflip} & \emojiWord{assets/comet3.pdf} & tok/s & \textbf{H} & \textbf{P} & \textbf{\pflip} & \emojiWord{assets/comet3.pdf} & tok/s & \textbf{H} & \textbf{P} & \textbf{\pflip} & \emojiWord{assets/comet3.pdf} & tok/s \\
        \midrule
        ZS      & 0.21 & 0.10 & 0.05 & 0.69 & 39.8 & 0.15 & 0.08 & 0.04 & 0.71 & \textbf{25.7} & 0.24 & 0.08 & 0.05 & 0.70 & \textbf{25.9} \\
        ZS-Exp.   & 0.30 & 0.22 & 0.16 & 0.68 & \textbf{41.3} & 0.41 & 0.22 & 0.18 & 0.72 & 24.6 & 0.56 & 0.23 & 0.21 & 0.69 & 25.5 \\
        MS      & 0.37 & 0.24 & 0.16 & 0.69 & 36.0 & \textbf{0.48} & 0.31 & 0.27 & \textbf{0.73} & 16.8 & 0.58 & 0.32 & \textbf{0.28} & \textbf{0.73} & 17.2 \\
        \midrule
        ActAdd  & 0.27 & 0.22 & 0.12 & 0.67 & 40.2 & 0.32 & 0.24 & 0.20 & 0.70 & 25.3 & 0.55 & 0.36 & \textbf{0.28} & 0.70 & 24.2 \\
        ReFT    & 0.31 & 0.22 & 0.18 & \textbf{0.70} & 40.7 & 0.46 & 0.34 & 0.27 & 0.67 & 25.5 & 0.53 & \textbf{0.38} & 0.26 & 0.70 & 24.7 \\
        SAE Cont. & \textbf{0.39} & \textbf{0.27} & \textbf{0.19} & \textbf{0.70} & 41.1 & 0.46 & 0.33 & \textbf{0.29} & 0.72 & 25.1 & \textbf{0.59} & 0.31 & 0.27 & 0.72 & 23.1 \\
    \bottomrule[1.5pt]
    \end{tabular}
    }
    \caption{Averaged metric scores across all tested languages (per-language breakdown in \Cref{app:all-models}). \textbf{H}: human style accuracy, i.e. $p($\Hone$) + p($\Htwo$)$. \textbf{P}: personalization accuracy $p(H^+)$ for the desired style. \textbf{\pflip}: \% of segments for which style conditioning flips the classifier prediction from \MT~to $H^+$. 
    $\alpha = 5$ is used for SAE Cont. results.}
    \label{tab:results-averaged}
    \vspace{-8pt}
\end{table*}

\paragraph{Metrics}
We evaluate our approaches on a held-out test set sourced from the \textsc{par3} dataset. For assessing style accuracy, we use the \classifier{}s described in~\Cref{sec:discern}. We define three submetrics employing the classifier probability distribution over the three classes (\MT, \Hone, \Htwo) to better analyze different aspects of model outputs. First, we define the \textbf{H} accuracy as the \classifier{}'s total probability assigned to human-like translations, $p($\Hone$)~+~p($\Htwo$)$, thereby measuring generic \textit{human-like} features of the text. We instead define \textbf{P} accuracy as the probability of the desired $H^+$ class, which corresponds to specific stylistic traits of a human translator. Finally, we use \textbf{\pflip} to indicate the proportion of examples for which the \classifier{} prediction flips from \MT~ to $H^+$ after prompting or steering is applied, pinpointing examples for which the method has an observable effect on generation style. To ensure that our interventions do not result in a degradation of overall translation quality, we also employ \comet{}\footnote{\href{https://huggingface.co/Unbabel/wmt22-comet-da}{Unbabel/wmt22-comet-da}}~\citep{rei-etal-2020-comet} using the $H^+$ as reference.



\begin{figure}
    \centering
    \includegraphics[width=\linewidth]{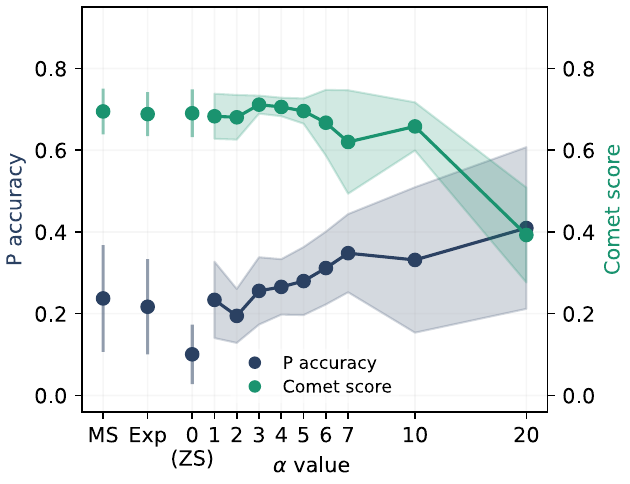}
    \includegraphics[width=0.9\linewidth]{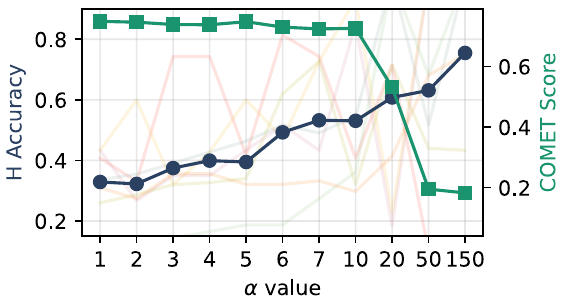}
    \caption{Effect of various steering intensity $\alpha$ on style accuracy and translation quality for Gemma 2 2B. \textbf{Top:} \textbf{P} accuracy for SAE Cont. and prompting baselines (MS, ZS-Exp and ZS). \textbf{Bottom:} H accuracy for high $\alpha$ showing a steep drop in translation quality while style accuracy increases.}
    \label{fig:alpha-sel}
    \vspace*{-12pt}
\end{figure}

\paragraph{The $\alpha$ trade-off}
We begin by verifying the optimal steering intensity $\alpha$ for our SAE steering technique. We primarily focus on results from Gemma 2 2B, for which we ran a comprehensive sweep over all relevant hyperparameters.\footnote{Larger models were evaluated using a subset of the best-performing configurations. Details in Appendix~\ref{app:all-models}.} Figure~\ref{fig:alpha-sel} (top) shows the influence of $\alpha$ on MT personalization accuracy and fluency, averaged across all translators and all tested languages. For values of $\alpha \leq 3$, performance remains close to that of the MS baseline, indicating that the contrastive method is effectively isolating latents associated with the desired style. As $\alpha$ increases, performance generally exceeds the MS approach, achieving greater stylistic accuracy with minimal impact on translation quality. However, for $\alpha \geq 10$, we observe a degradation in translation fluency, represented by \comet{}.

Following the SAE steering setup of~\citet{ferrando2025iknowentityknowledge}, we also experiment with very high $\alpha$ (up to 150), obtaining \textbf{H} accuracies approaching 100\% for some languages but also resulting in abysmal translation quality, with generally nonsensical outputs (\Cref{fig:alpha-sel} bottom). These results suggest that SAE features identified by our contrastive approach correspond to human-like constructions, provided that their upweighting raises the \classifier{} confidence in predicting \Hone~or \Htwo, even when the resulting quality is clearly lacking.\footnote{A qualitative evaluation is provided in~\Cref{app:output-examples}.} We leave an exploration on using such examples to further improve style classifiers to future work. Ultimately, we identify $\alpha = 5$ as an appropriate steering intensity that balances personalization and fluency, and employ it in our main evaluation.

\subsection{Results and Discussion}
\label{sec:results}

Table \ref{tab:results-averaged} presents the performance of tested models across prompting and steering setups, averaged across all languages and styles, i.e., using \Hone~and \Htwo~as $H^+$ for each language. We find that contrastive SAE steering (SAE Cont.) yields a good balance between style accuracy and translation quality, with results comparable to the strong MS prompting baseline. Notably, SAE Cont. outperforms all other tested methods for the smaller Gemma 2 2B model. We conjecture this could be due to the larger models' ability to incorporate in-context information more effectively, reducing the benefits of ad-hoc interventions.

\begin{figure}
    \centering
    \includegraphics[width=0.9\linewidth]{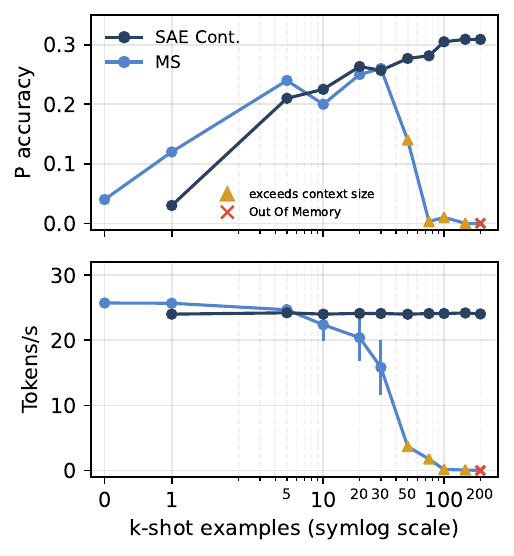}
    \caption{Personalization accuracy \textbf{P} (top) and inference speed (Tokens/s) (bottom) across in-context demonstration counts, using Gemma-2 9B for Russian $\rightarrow$ English translation. More results in \Cref{app:full-ICL}.}
    \label{fig:ru-max-icl}
    \vspace{-12pt}
\end{figure}

\paragraph{Is steering computationally efficient?} To quantify the impact of various steering strategies on inference speed, for each method in \Cref{tab:results-averaged} we include the resulting generation speed in tokens per second (tok/s) estimated over the whole test set. Despite their comparable performances, we observe that MS prompting is systematically slower than SAE Cont. steering across all tested models when using 20 in-context demonstrations. We set out to test the trade-off between style conditioning and computational efficiency, focusing on the translation direction with more paragraphs available (Russian$ \rightarrow$English) as we vary the number of examples used in prompting and steering. Our results in \Cref{fig:ru-max-icl} show that increasing the number of examples used to identify features yields consistent improvements for SAE Cont., while scaling in MS prompting is limited by the model's limited context size. Moreover, the inference-time cost of SAE Cont. steering remains constant when increasing the demonstrations used for selecting steered features, while MS prompting undergoes a significant slowdown when demonstrations are added to the prompt. Overall, these results strongly highlight how the SAE Cont. method, while comparable to MS prompting in smaller setups, can result in more scalable performances without requiring additional inference time computations. We report additional evidence in~\Cref{app:full-ICL}.


\paragraph{Is SAE steering an ``internalized'' MS prompting?} In light of the comparable results obtained by SAE steering and MS prompting, we set out to investigate whether the two methods also modify model activations in a similar way. For this purpose, we use the \textit{probing classifiers} introduced in~\Cref{sec:probing}, which were trained on MS-conditioned last prompt token activations for predicting the style of the resulting text. We categorize test examples in three groups: one in which steering flips the original classifier prediction to a human label, i.e. \Hone~or \Htwo~(\MT~$\rightarrow H$), and two in which prediction remains unchanged (\MT~$\rightarrow$\MT, $H\rightarrow H$). For each example, we collect steered activations $z_{\text{steer}}$ using the SAE Cont. method and calculate the average probability of human-like output ($H$) predicted by the probe over each category. Our results in \Cref{tab:prober-steer} show that, for both Gemma models, probes can predict successful (\MT~$\rightarrow H$) and unsuccessful (\MT~$\rightarrow$\MT) steering outcomes from SAE-steered activations, despite being trained only on MS prompting examples. In settings where the original output already matches human style ($H \rightarrow H$), instead, probes reach lower accuracy with broader confidence intervals, denoting higher uncertainty. The ability of probes to generalize to SAE steering from prompting examples provides strong evidence that \textbf{SAE Cont. steering mirrors the effect of in-context demonstrations from MS prompting on model activations}. This suggests, in turn, that SAE features identified by our contrastive setup might act as efficient ``internalized summaries'' of prompt demonstrations.

\begin{table}[]
    \small
    \centering
    \begin{tabular}{lccc}
    \toprule
               & \MT~$\rightarrow H$ & \MT~$\rightarrow$ \MT & $ H \rightarrow  H$  \\
               \midrule
    Gemma 2 2B & 0.94 \scalebox{0.8}{$\pm0.01$}  & 0.07 \scalebox{0.8}{$\pm0.02$}  & 0.72 \scalebox{0.8}{$\pm0.15$} \\
    Gemma 2 9B & 0.93 \scalebox{0.8}{$\pm0.02$}  & 0.12 \scalebox{0.8}{$\pm0.10$}  & 0.68 \scalebox{0.8}{$\pm0.19$} \\ 
    \bottomrule
    \end{tabular}
    \caption{Estimated $p(H)$ for SAE-steered activations using \Cref{sec:probing} binary probes trained on MS-conditioned activations, across three settings identified by \classifier{} input/output labels.}
    \label{tab:prober-steer}
    \vspace{-12pt}
\end{table}

\section{Conclusion and future work}
We conducted a broad evaluation of various prompting and steering approaches for personalizing LLM-generated translations. Our evaluation targets a practical, real-world application of literary translation, addressing the underexplored challenge of steering LLM generations in a linguistically rich and stylistically sensitive domain.
Through comprehensive evaluation across multiple languages, novels, and models, we demonstrate that our proposed SAE-based approach outperforms prompting and alternative steering techniques, resulting in a good balance of style conditioning and translation quality at negligible inference-time cost. In particular, our proposed method scales effectively to large amounts of examples, bypassing the context limitations and slowdowns of in-context demonstrations.

Although faithfully replicating individual human translation styles remains a highly challenging task, our approach achieves strong alignment with human translation quality and translator-specific style features. In particular, the effectiveness of our proposed approach on smaller models might enable MT personalization in smaller-scale computational settings, easing further research on how style information is encoded and produced by language models.


\section*{Limitations}

While our work demonstrates the potential of steering LLMs for MT personalization using sparse autoencoders, we acknowledge some limitations in our setup.

Firstly, the generalizability of our findings is constrained by the scope of our experiments. We focused on literary translation from seven specific source languages into English and evaluated three LLMs of relatively small size. Consequently, the effectiveness of SAE-based steering and the identified optimal layers for intervention may not directly transfer to other language pairs, significantly different model architectures or sizes, or distinct domains beyond literary texts. Further research is needed to assess the robustness of our approach across a broader range of linguistic and modeling contexts.

Secondly, the computational overhead associated with training sparse autoencoders presents a practical limitation to the widespread adoption of our proposed method. Although many open-source pre-trained SAEs are now available, training new SAEs from scratch may be required for newer models, which might pose a problem for researchers with limited computational resources. The current availability of pre-trained SAEs also restricts model choice, as seen with the Llama 3.1 8B model where the SAE we selected was available for a single, potentially suboptimal layer.

Finally, our investigation primarily focused on downstream performance and the impact of various personalization strategies on model activations. However, we did not pursue a mechanistic understanding of the "personalization circuits" within the LLMs. Future work could adopt a more fine-grained, mechanistic interpretability approach to study how specific SAE latents or combinations thereof encode and manipulate nuanced stylistic features, thereby providing deeper insights into the underlying processes of LLM personalization.


\bibliography{anthology,custom}
\bibliographystyle{misc/acl_natbib}

\clearpage
\appendix





\section{Experimental Reproducibility}

In this section, we provide every parameter we use for the reproducibility of our experiments setups.

\subsection{Base prompt}

We use the same prompt template across all methods: ZS (which corresponds to the original model translation), ZS-Exp. (detailed in~\Cref{app:explain}), MS, ActAdd, ReFT, and SAE Cont. This prompt, shown in Listing~\ref{lst:explain-final}, instructs the model to translate the source sentence while explicitly preventing it from adding any explanations about the translation process. Since all test models are Instruction Tuned, we utilize their native chat templates to preprocess the input accordingly. For multi-shot examples, the \textit{user} and \textit{assistant} turns are repeated for each example, always using the same prompt structure.

\subsection{Classifier training}
\label{app:classifiers}
All classifiers are fine-tuned from the \texttt{xlm-roberta-large} model\footnote{\href{https://huggingface.co/FacebookAI/xlm-roberta-large}{FacebookAI/xlm-roberta-large}}, using a linear classification head. Training is conducted for 6 epochs with a learning rate of 2e-5 and a batch size of 32, selecting the best model checkpoint based on validation accuracy.

Training data only includes generations from models and the translator without any source text. It is also perfectly balanced, as each paragraph provides one instance for all three labels: \Hone, \Htwo, and \MT. The total size of the training set varies depending on the number of paragraphs in the chosen novel. On average, we obtain approximately 830 instances, resulting in a total of around 2,490 labeled examples for training (see~\ref{tab:novels-details}). Validation and test sets are strictly held out and never seen during training. Additionally, they do not include the small 20-example subsets used in the MS, ZS-Exp. and SAE Cont. setups.

\subsection{ReFT training}
\label{app:reft-training}
ReFT training was conducted using the PyReFT toolkit from the original authors\footnote{\href{https://github.com/stanfordnlp/pyreft}{\texttt{stanfordnlp/pyreft}}}. We applied the intervention at the same hook point used by other steering methods - specifically, the layer output corresponding to the residual stream at the selected layer. The training configuration includes a \texttt{low rank dimension} of 4, \texttt{lora alpha} set to 32, and a \texttt{lora dropout} of 0.05. ReFT was trained on the same 20 prompts used in the MS setup, for a total of 100 epochs.

\subsection{SAE Cont.}
We use the NNsight library~\citep{fiottokaufman2024nnsightndifdemocratizingaccess} to extract and manipulate model activations for all steering experiments. The source code is publicly available in the repository linked in the main body of this paper. For consistency, we use the same set of contrastive examples employed in the MS approach.

Algorithm~\ref{alg:latent-steering} outlines the procedure for latent-based steering. It enhances features identified as relevant to personalization while simultaneously suppressing those negatively correlated with the task.

\begin{algorithm}
\small
\caption{Contrastive SAE Steering}
\KwIn{Input activation $z$, SAE model \texttt{sae}, target latents expected value $\mathbb{E}^+[X_i]$, contrast latents expected value $\mathbb{E}^-[X_i]$, steering coefficient $\alpha$}
\KwOut{Steered activation $z_{\text{new}}$}

$x \leftarrow \texttt{sae.encode}(z)$\;
$m \leftarrow \text{length}(x)$\;
\For{$i \leftarrow 1$ \KwTo $m$}{
    \If{$\mathbb{E}^+[X_i] > x[i]$}{
        $x[i] \leftarrow \mathbb{E}^+[X_i]$
    }
    \If{$\mathbb{E}^-[X_i] < x[i]$}{
        $x[i] \leftarrow \mathbb{E}^-[X_i]$
    }
}
$z_{\text{steer}} \leftarrow \alpha \cdot \texttt{sae.decode}(x)$\;
\Return{$z_{\text{steer}}$}
\label{alg:latent-steering}
\end{algorithm}

\subsection{ZS-Exp.}
\label{app:explain}
For ZS-Exp., we used GPT-4o (June 2025) to generate explanations detailing the stylistic differences between a base translation and a target human translation. The prompt template used for this task is shown in Listing~\ref{lst:explain-template}, using the same 20 examples as in the MS and SAE Cont. setups. Example outputs for different novels are shown in Listing~\ref{lst:explain-example}. Finally, these generated guidelines are used to prompt the evaluated models, following the template shown in Listing~\ref{lst:explain-final}.

\begin{figure*}[h]
    \centering
    \small
    \begin{minipage}{0.9\textwidth}
    \begin{lstlisting}[style=fancyquote, caption={Prompt template used to get GPT 4o explanation using translation examples.}, label={lst:explain-template}]

Objective: Identify stylistic choices in translations for personalization purposes.

You will be provided with a source text, a standard translation, and a target translation by a specific translator whose style we want to emulate.
Your task is to analyze the 'Target translation' by comparing it to the 'Base translation' and the 'Source text'.
Identify and list the distinctive stylistic patterns, choices, and preferences exhibited in the Target translation.

These stylistic cues should help another translator (or an AI) to adapt their translations to match the style of the target translator.

Source text: @<source text here>@
Base translation: @<MT  text here>@
Target translation: @<H+ translation here>@

@<... repeat Source, MT and Target>@

Please extract a concise list of key stylistic cues. Focus on aspects such as vocabulary choices, sentence structure, tone and register, handling of cultural nuances, punctuation/formatting preferences and overall creativity.

Output a short list of stylistic cues as bullet points. Write the list as if you were directly giving the guidelines to the translator and avoid using specific examples.
    \end{lstlisting}
    \end{minipage}
\end{figure*}

\begin{figure*}[h]
    \centering
    \small
    \begin{minipage}{0.9\textwidth}
    \begin{lstlisting}[style=fancyquote, caption={Examples of explanation obtained from GPT 4o when comparing different translations from different novels.}, label={lst:explain-example}]

@When comparing H1 and MT for Beware of Pity (German):@
    - Maintain a tone that is professional, thoughtful, and subtly persuasive.
    - Avoid overly technical jargon unless necessary; explain specialized terms briefly if used.
    - Preserve the author's voice, keeping a balance between academic rigor and narrative engagement.
    - Ensure smooth transitions between sentences and paragraphs to support coherent argumentation.
    - Translate idiomatic expressions in a way that retains their intended effect, even if the wording differs.

@When comparing H1 and H2 for Beware of Pity (German):@
    - Keep the tone warm, welcoming, and direct - avoid overly formal or distant language.
    - Use active voice wherever possible to maintain energy and engagement.
    - Avoid idioms or expressions that may not translate culturally; aim for universal accessibility.
    - Maintain consistent tone and register throughout, adapting to the intended audience's familiarity with the subject.
    - Respect the rhythm and structure of the original, but feel free to adjust sentence length for readability.

@When comparing H1 and H2 for Crime and Punishment (Russian):@
    - Prefer dynamic over formal vocabulary: Opt for vivid or emotionally charged words when available.
    - Add tonal nuance and emotional shading: Enrich dialogues and narration with subtle shifts in tone, especially sarcasm, understatement, or irony, to match character voice or mood.
    - Use contractions and familiar phrasing: Employ contractions and relaxed expressions to preserve spoken character.
    - Expand or rephrase for clarity and voice: Don't hesitate to slightly reword or elaborate if it strengthens tone, clarifies intent, or enhances character differentiation.
    - Favor rhythmic, flowing sentence structure: Break long, formal sentences into multiple shorter clauses or use punctuation (dashes, ellipses) for dramatic or emotional effect.
    - Reflect subtle character dynamics: Infuse lines with interpersonal undertones (like defiance, deference, or sarcasm) that may not be explicit in the original.
    - Preserve or recreate emotional tension: Use word choice and pacing to sustain psychological nuance, unease, or irony.
    - Use expressive punctuation and formatting: Favor dashes, ellipses, and italic like emphasis (through word placement) to reflect emotional cadence or interruptions.

    \end{lstlisting}
    \end{minipage}
\end{figure*}

\begin{figure*}[h]
    \centering
    \small
    \begin{minipage}{0.9\textwidth}
    \begin{lstlisting}[style=fancyquote, caption={Zero shot template template when prompting language models with different setups}, label={lst:explain-final}]

Translate the following sentence between the angular parentheses into English.
@if setup == ZS-Exp@
@{@
    Follow the following guidelines when translating:
    @<explanations here>@
@}@

The original sentence is: @<source text>@.

Remember to write only the translation, without any additional text or explanation.
    \end{lstlisting}
    \end{minipage}
\end{figure*}

\section{Results across languages}
\label{app:all-models}

\subsection{Classifiers}
\label{app:classifier-results}

We show in Table~\ref{tab:classifiers-perf} results for every \classifier{} trained for each model and for each language.

\begin{table}
    \small
    \centering
    \begin{tabular}{l|ccc}
    \toprule[1.5pt]
    \textbf{Lang.} & \textbf{Gemma 2 2B} & \textbf{Gemma 2 9B} & \textbf{Llama 3.1 8B} \\
    \midrule
    DE & 0.89 & 0.90 & 0.84 \\
    RU & 0.92 & 0.90 & 0.91 \\
    ZH & 0.99 & 0.98 & 0.98 \\
    IT & 0.78 & 0.85 & 0.80 \\
    NL & 0.79 & 0.78 & 0.82 \\
    FR & 0.88 & 0.87 & 0.90 \\
    JA & 0.76 & 0.79 & 0.76 \\

    \bottomrule[1.5pt]
    \end{tabular}
    \caption{Accuracy of model- and language-specific 3-way (\MT, \Hone, \Htwo) classifiers  on balanced held-out sets for every language. Random baseline: $0.33$.}
    \label{tab:classifiers-perf}
    \vspace{-8pt}
\end{table}


\subsection{Prompting and steering results}
\begin{figure*}
    \centering
    \includegraphics[width=\linewidth]{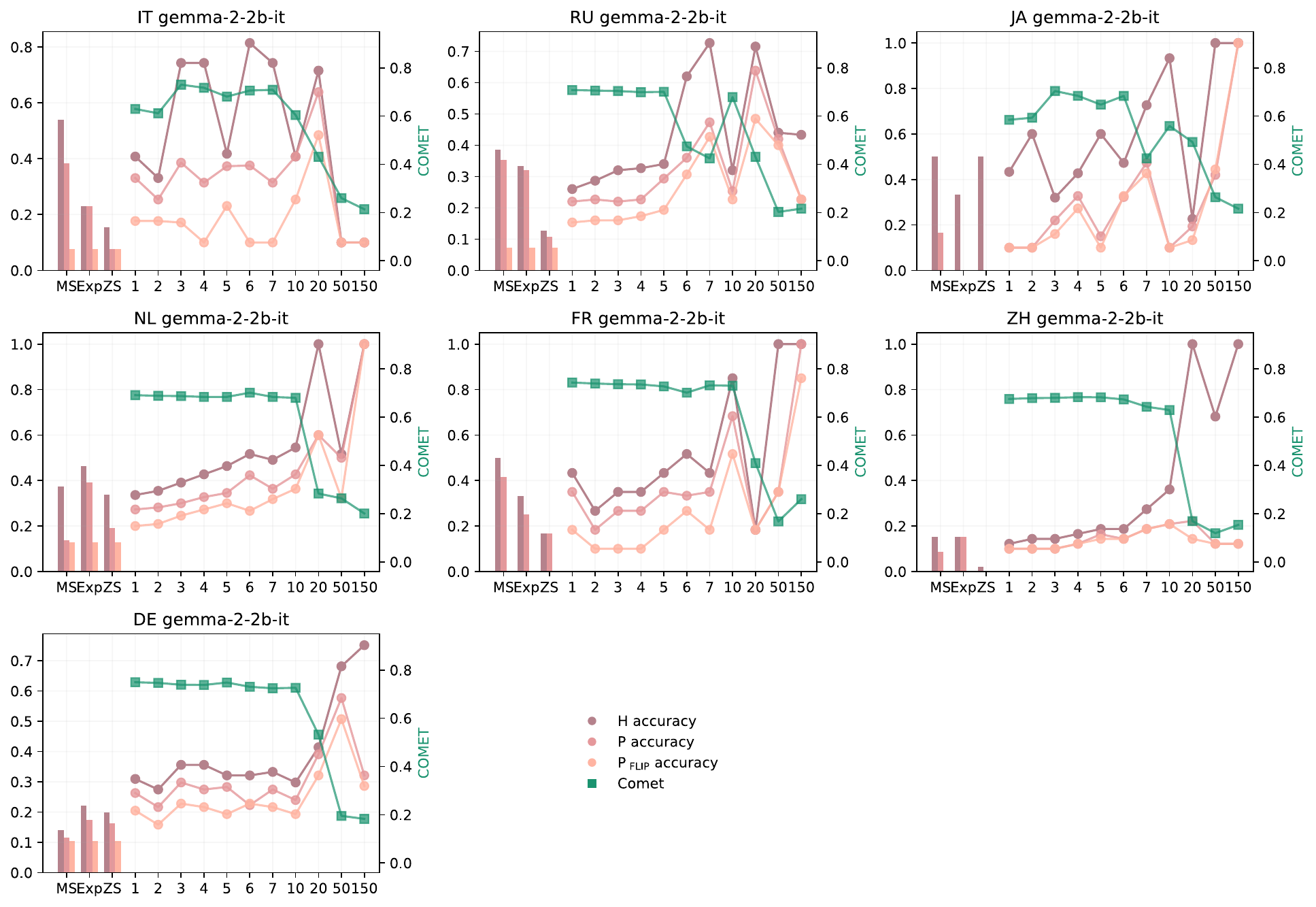}
    \caption{Results for every language on Gemma 2 2B.}
    \label{fig:gemma2b-all}
\end{figure*}

\begin{figure*}
    \centering
    \includegraphics[width=\linewidth]{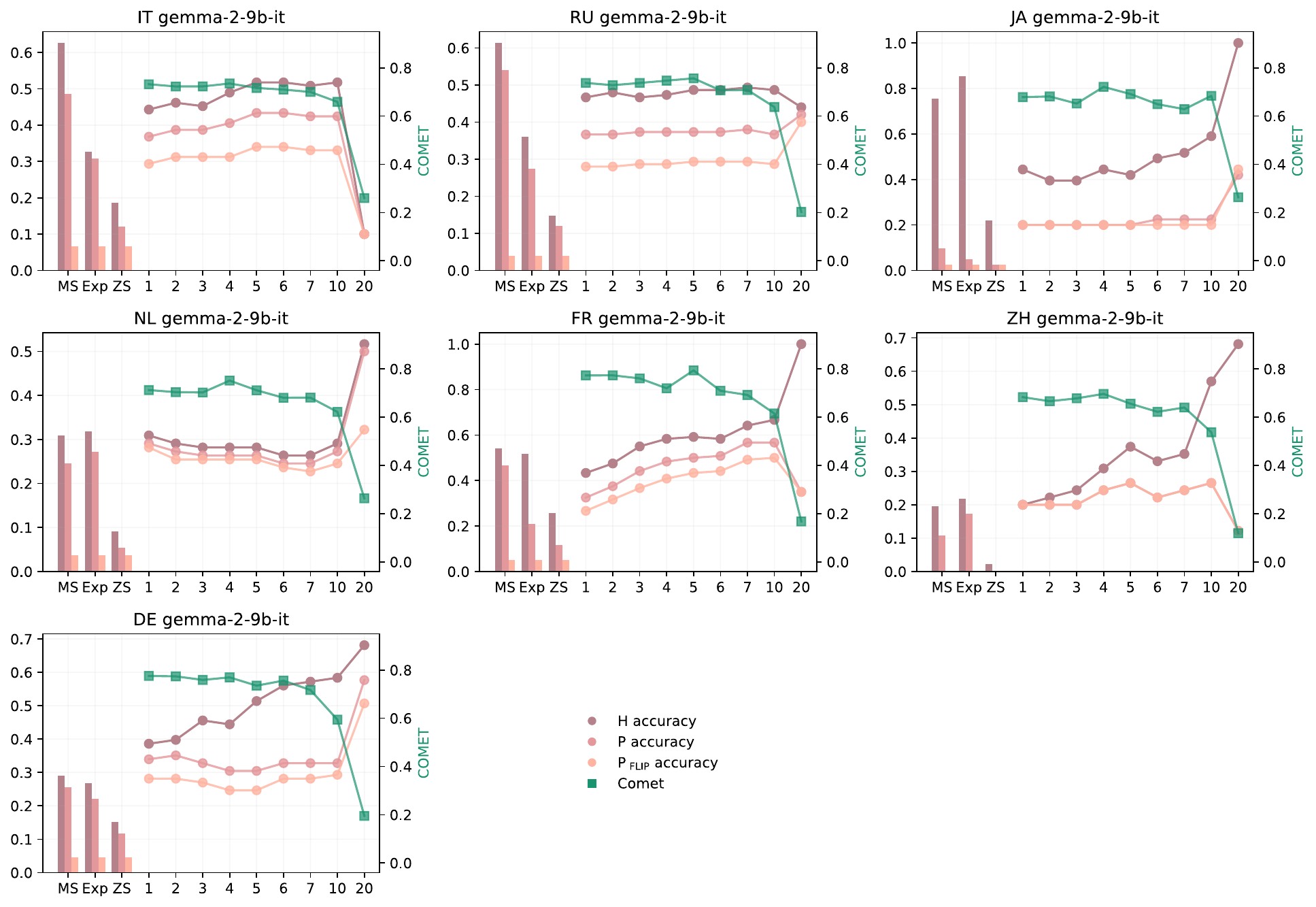}
    \caption{Results for every language on Gemma 2 9B.}
    \label{fig:gemma9b-all}
\end{figure*}

\begin{figure*}
    \centering
    \includegraphics[width=\linewidth]{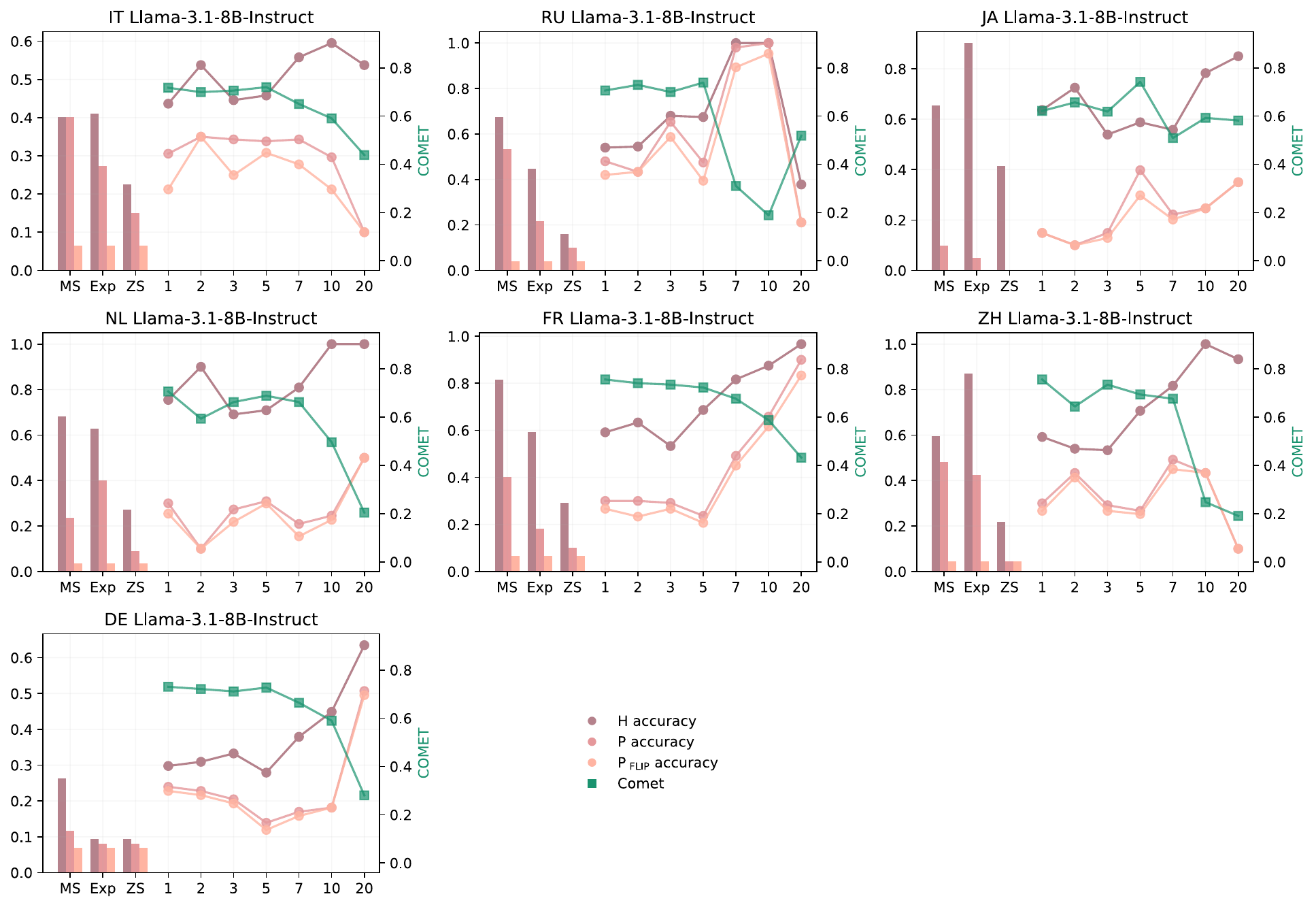}
    \caption{Results for every language on Llama 3.1 8B.}
    \label{fig:llama8b-all}
\end{figure*}

We present detailed plots of the results for each novel across the three evaluated models in Figure~\ref{fig:gemma2b-all} (Gemma 2 2B), Figure~\ref{fig:gemma9b-all} (Gemma 2 9B), and Figure~\ref{fig:llama8b-all} (Llama 3.1 8B). These plots display the performance of all evaluated methods, reporting the three submetrics: \textbf{H} accuracy (general human-likeness), \textbf{P} accuracy (translator-specific accuracy), and \textbf{\pflip} (personalized flip accuracy), alongside the corresponding \comet{} scores measuring translation quality.

\section{Examples from Dataset and Conditioning Outputs}
\label{app:output-examples}
We present in Tables~\ref{tab:examples-all-zh} and~\ref{tab:examples-all-it} a selection of examples from two different languages, showcasing outputs from each of the tested setups. For each example, we also report the corresponding classification label predicted by the \classifier{} and the associated \comet{} score. Additionly table~\ref{tab:extreme-alphas} shows some examples of models generating output aligned with the Human translator according to the \classifier{} but with a low \comet{} score corresponding to an almost unreadable output due to extreme $\alpha$ values.

\section{Additional Experiment on Contrastive Setups to Isolate Style Features}


To isolate features relevant to a specific translation style, we adopt a contrastive approach between different parallel translation examples. Specifically we test two setups:
\begin{itemize}
    \item $M \leftrightarrow H$: contrasting the target style we aim to achieve ($H^+$) with a "clean" style derived from the model's original machine translation ($M$);
    \item $H_\alpha\leftrightarrow H_\beta$ contrasting the desired style with the style of another human translator.
\end{itemize}

The $M \leftrightarrow H$ setup is designed to extract features that distinguish the target human-like style from a standard, model-generated translation. The $H_\alpha\leftrightarrow H_\beta$ setup, on the other hand, aims to isolate the features that differentiate one translator's style from another. While the latter more closely corresponds to the notion of translation style (as a property unique to each translator), we chose to experiment with both since prior works have shown that contrasting the desired behavior with the model's baseline behavior tends to yield the most effective results when steering LLMs.

\paragraph{Which contrastive setup is better?}
We present in Table~\ref{tab:ht-pt} the results for both contrastive setups. For both the ZS-Exp. and SAE Cont. approaches, the results are generally comparable, making it difficult to identify a clear best-performing method. However, larger models occasionally achieve higher performance, which we hypothesize may stem from their enhanced ability to disentangle personalization-relevant features even without explicit guidance.


\section{Additional Results when Scaling Demonstrations Count}
\label{app:full-ICL}
\begin{figure*}
    \centering
    \includegraphics[width=\linewidth]{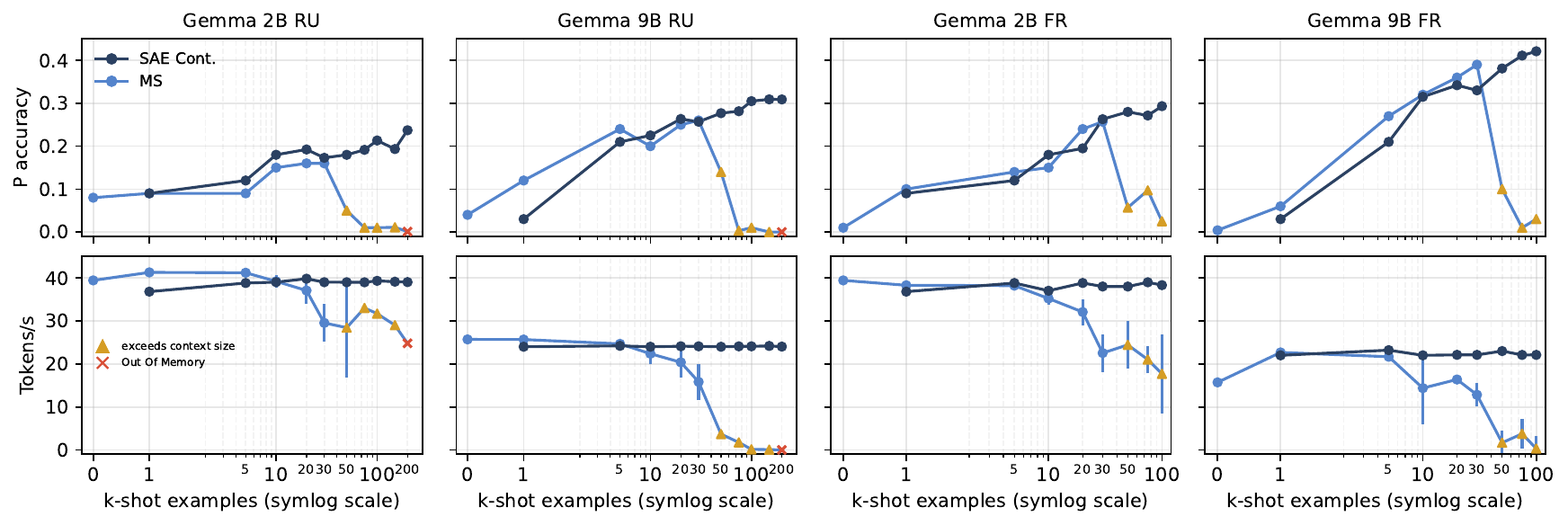}
    \caption{Complete results when comparing the MS approach to our \contht for the Gemma models (2B and 9B) on the largest novels, evaluated at the paragraph level, in Russian (RU) and French (FR).}
    \label{fig:full-icl}
\end{figure*}

\Cref{fig:full-icl} presents the full set of experiments, including two languages (Russian and French) and two models (Gemma 2 2B and 9B), complementing the results of \Cref{fig:ru-max-icl}. The two languages were selected as they presented the largest collection of translated paragraphs available (see~\Cref{tab:data-examples}), allowing us to scale the ICL set without substantially reducing the size of the training data or impairing classifier performance. Notably, as also shown in~\Cref{fig:ru-max-icl} the personalization accuracy \textbf{P} of our contrastive approach improves across all models and language tested, even surpassing the multi-shot (MS) baseline. In all cases, the MS baseline reaches the LLM context size limit (around ICL $\geq 50$), after which the performance degrades significantly.
Similarly, the tok/s efficiency metric highlights the advantage of our contrastive approach: It brings only a minimal, roughly linear, decrease in inference-time speed (compared to ZS, i.e. ICL $=0$), whereas the MS setup exhibits a progressively larger computational impact as the number of in-context examples increases.

\begin{table}[t]
    \small
    \centering
    \begin{tabular}{l ccccc}
    \toprule[1.5pt]
        \textbf{Method} & \textbf{H} & \textbf{P} & \textbf{\pflip} & \emojiWord{assets/comet3.pdf} & tok/s \\
        \midrule
        \multicolumn{6}{c}{\textbf{Gemma 2 2B}} \\ 
        \midrule
        \exht   & 0.30 & 0.22 & 0.16 & 0.68 & \textbf{41.3} \\
        \expt   & --   & 0.20 & 0.14 & 0.69 & 41.2          \\
        \cmidrule(lr){2-6}
        \contht & \textbf{0.39} & \textbf{0.27} & \textbf{0.19} & \textbf{0.70} & 41.1 \\
        \contpt &   --   & \textbf{0.27} & 0.18 & 0.69 & 40.8  \\
        \midrule
        \multicolumn{6}{c}{\textbf{Gemma 2 9B}} \\
        \midrule
        \exht   & 0.41 & 0.22 & 0.18 & 0.72 & 24.6 \\
        \expt   &  --   & 0.23 & 0.19 & \textbf{0.73} & \textbf{24.4} \\
        \cmidrule(lr){2-6}
        \contht & \textbf{0.46} & 0.33 & \textbf{0.29} & 0.72 & 25.1 \\
        \contpt & -- & \textbf{0.35} & \textbf{0.29} & \textbf{0.73} & 24.9 \\
        \midrule
        \multicolumn{6}{c}{\textbf{LLaMA 3.1 8B}} \\
        \midrule
        \exht   & 0.56 & 0.23 & 0.21 & 0.69 & 25.5 \\
        \expt   & -- & 0.30 & 0.26 & 0.70 & 25.5 \\
        \cmidrule(lr){2-6}
        \contht & \textbf{0.59} & 0.31 & 0.27 & \textbf{0.72} & \textbf{23.1} \\
        \contpt &   -- & \textbf{0.33} & \textbf{0.28} & \textbf{0.72} & 23.6  \\        
    \bottomrule[1.5pt]
    \end{tabular}
    \caption{Averaged metric scores across all tested languages for the $M \leftrightarrow H$ and $H_\alpha \leftrightarrow H_\beta$ variants of ZS-Exp. and SAE Cont. methods. \textbf{H}: human style accuracy, i.e. $p($\Hone$) + p($\Htwo$)$. \textbf{P}: personalization accuracy $p(H_x)$ for the target style. \textbf{\pflip}: Proportion of segments for which steering has a causal impact on personalization. 
    $\alpha = 5$ is used for SAE Cont. results.}
    \label{tab:ht-pt}
    \vspace{-8pt}
\end{table}

\begin{table*}
    \small
    \centering
    \scalebox{0.9}{
\begin{tabular}{p{0.12\textwidth}p{0.70\textwidth}C{0.1\textwidth}C{0.08\textwidth}}
\toprule[1.5pt] 
Source (ZH) & 遂又往後看時，只見畫著一張弓，弓上掛著香櫞。也有一首歌詞云：「二十年來辨是非，榴花開處照宮闈。三春爭及初春景，虎兕相逢大夢歸。」& \classifier{} & \comet{} \\
\midrule
\Hone & \textit{Forthwith, therefore, he prosecuted a further perusal of what came next, when he caught sight of a picture of a bow. On this bow hung a citron. There was also this ode: Full twenty years right and wrong to expound will be thy fate! What place pomegranate blossoms come in bloom will face the Palace Gate! The third portion of spring, of the first spring in beauty short will fall! When tiger meets with hare thou wilt return to sleep perennial.} & \Hone & - \\
\Htwo & \textit{The picture that followed was of a bow with a citron hanging from it, followed by what looked like the words of a song: You shall, when twenty years in life's hard school are done, In pomegranate-time to palace halls ascend. Though three springs never could with your first spring compare, When hare meets tiger your great dream shall end.} & \Htwo & - \\
\midrule
\textcolor{pastelorange}{Gemma 2 2B} & \textit{When he turned back to look, he saw a bow hanging on it, with a plum blossom. There was also a poem that said: "For twenty years, I have been trying to distinguish right from wrong, and the plum blossoms bloom in the palace. In the spring, the tigers and the rhinoceroses meet, and they dream of returning home."} & \MT & 0.49 \\[4pt]
\textcolor{pastelorange}{Gemma 2 9B} & \textit{Looking back, he saw a painting of a bow with a quiver hanging on it. There was also a line of lyrics that said: \"For twenty years, I have distinguished right from wrong, and the blossoming of the pomegranate flowers illuminates the imperial court. Three springs have passed, and now it is the beginning of spring again, like a dream where tigers and rhinoceroses meet.} & \MT & 0.56 \\[4pt]
\textcolor{pastelorange}{Llama 3.1 8B} & \textit{As I looked back, I saw a painting of a bow with a censer on it. There was also a poem that said: \"Twenty years of distinguishing right from wrong, the lotus flower blooms in the palace. Three springs competing with the scenery of the first spring, the tiger and the rhinoceros meet and return to their dream."} & \MT & 0.53 \\[4pt]
\midrule
\exht & \textit{And as he turned his gaze back upon the scene once more, there was displayed an archer's bow strung tautly; upon its surface hung a fragrant plum blossom.  A poem also recited thus spake: "For twenty years I have sought truth amidst strife, where blooming durian flowers illuminate the palace walls. In spring's embrace, they vie for beauty, tigers clashing in dreams of return."} & \MT & 0.60 \\
\expt & \textit{And as he turned his gaze back upon the scene once more, there was displayed an archer's bow strung tautly; upon its surface hung a fragrant plum blossom. A poem also recited thus:
"For twenty years I have sought truth amidst strife, where blooming durian flowers illuminate the palace walls. In spring's embrace, they vie for beauty, tigers clashing in dreams of return."} & \MT & 0.61 \\
MS & \textit{Then, turning around, he noticed a picture depicting a bow hanging on its string, adorned with incense sticks.  A verse read: "For twenty years I've tried to discern truth from falsehood, and wherever fragrant plum blossoms bloom, the palace gates shine brightly."} & \Hone & 0.65 \\
\midrule
ActAdd & \textit{When he turned back to look, he saw a bow hanging on it, with a plum blossom. There was also a poem that said: "For twenty years, I have been trying to tell right from wrong, and the plum blossoms bloom in the palace. In spring, the tigers and rhinoceroses meet, and they dream of going back home."} & \MT & 0.53 \\
ReFT & \textit{When he turned back to look, he saw a bow hanging there, adorned with a plum blossom. There was also a poem that said: "For twenty years, I have tried to distinguish right from wrong, and plum blossoms bloom within the palace. In spring, tiger and rhinoceros meet, dreaming of their return home."} & \MT & 0.58 \\
\contht \\~~ $\alpha = 1$ & \textit{When he turned back to look, he saw a bow hung upon it, with a plum blossom.  There was also a verse which read: "For twenty years, I have struggled to discern truth from falsehood, and the plums blossomed where the imperial court resided. During the three springs, the tiger and rhinoceros met, dreaming of their return."} & \MT & 0.62 \\
~~ $\alpha = 5$ & \textit{He looked back and there hung a bow adorned with a plum blossom. A verse read: “Twenty years I’ve struggled for truth, where fragrant plums bloom near the imperial court. Spring arrives, tiger and rhino clash, dreaming of their return.} & \MT & 0.62 \\
~~  $\alpha = 150$ & \textit{The when the veryShellTruth lookWarmTheTheTheTigeratPoint [...]} & \MT & 0.11 \\
\contpt \\~~ $\alpha = 5$ & \textit{He looked back and saw a bow adorned with fragrant citron. A verse declared: "Twenty years I’ve weighed right and wrong; where pomegranate blossoms bloom, they light the palace gates. Spring’s full bloom cannot match early spring’s grace. When tiger meets rhino, one returns to the great dream."} & \Hone & 0.65 \\

\bottomrule[1.5pt]
\end{tabular}}
    \caption{Example paragraphs from ``Dream of the Red Chamber'' (ZH $\rightarrow$ EN) translated with \Hone personalization. Setup outputs are from Gemma 2 2B; only MS and \contpt successfully flip the \classifier{}' prediction.}
    \label{tab:examples-all-zh}
\end{table*}

\begin{table*}
    \small
    \centering
    \scalebox{0.9}{
\begin{tabular}{p{0.12\textwidth}p{0.70\textwidth}C{0.1\textwidth}C{0.08\textwidth}}
\toprule[1.5pt] 
Source (IT) & Intanto la stalla fu chiusa e Pinocchio rimase solo: e perch\'e erano molte ore che non aveva mangiato, cominci\'o a sbadigliare dal grande appetito. E, sbadigliando, spalancava una bocca che pareva un forno. & \classifier{} & \comet{} \\
\midrule
\Hone & \textit{When the stable door was closed, Pinocchio was left alone. And since he hadn't eaten in many hours, he began to yawn from hunger. And when he yawned, his mouth opened as wide as an oven.} & \Hone & - \\
\Htwo & \textit{The stable was then shut and Pinocchio was left alone. He had not eaten for many hours, and he began to yawn from hunger. And when he yawned he opened a mouth that seemed as wide as an oven.} & \Htwo & - \\
\midrule
\textcolor{pastelorange}{Gemma 2 2B} & \textit{At that moment the stable was closed and Pinocchio was left alone, and because it had been many hours since he had eaten, he began to yawn from a great hunger. And, yawning, he opened a mouth that seemed like a furnace.} & \MT & 0.79 \\[4pt]
\textcolor{pastelorange}{Gemma 2 9B} & \textit{Meanwhile, the stable was closed and Pinocchio was left alone: and because it had been many hours since he had eaten, he began to yawn from his great hunger. And, yawning, he opened a mouth that looked like an oven.} & \MT & 0.80 \\[4pt]
\textcolor{pastelorange}{Llama 3.1 8B} & \textit{Meanwhile the stable was closed and Pinocchio was left alone: and because it was many hours since he hadn't eaten, he began to yawn from great hunger. And, yawning, he opened a mouth that seemed like an oven.} & \MT & 0.80 \\[4pt]
\midrule
\exht & \textit{Meanwhile, the stable door was shut, leaving Pinocchio alone. His tummy rumbled because it had been hours since his last meal, and he started yawning widely from his hunger. With each yawn, his mouth opened like an oven. } & \MT & 0.78 \\
\expt & \textit{Meanwhile, the stable door was shut, leaving Pinocchio all alone.  His tummy rumbled because it had been hours since his last meal, and he started yawning widely from sheer hunger. With each yawn, his mouth gaped open like an oven. } & \Htwo & 0.80 \\
MS & \textit{Meanwhile, the stable door was shut and Pinocchio was left alone. Since it had been hours since he last ate, he began yawning from hunger.  Each yawn opened his mouth wide enough to look like an oven.} & \Htwo & 0.81 \\
\midrule
ActAdd & \textit{Meanwhile, the stable was closed and Pinocchio remained alone. Because many hours passed without eating, he started yawning from very big hunger. And while yawning, he opened his mouth wide like an oven.} & \MT & 0.75 \\
ReFT & \textit{Meanwhile, the stable was closed and Pinocchio stayed alone. Because many hours had passed since he last ate, he began to yawn from big hunger. And, yawning, he opened a mouth that looked like an oven.} & \MT & 0.77 \\
\contht \\~~ $\alpha = 1$ & \textit{Meanwhile, the stable door was shut and Pinocchio was left alone.  Because it had been many hours since he had eaten, he started to yawn with intense hunger. And, as he yawned, his mouth gaped open like an oven. } & \MT & 0.79 \\
~~ $\alpha = 5$ & \textit{Meanwhile, the stable door shut, leaving Pinocchio all alone.  Because it had been many hours since he’d eaten, he started to yawn with intense hunger. As he yawned, his mouth gaped open, looking like an oven.} & \Htwo & 0.82 \\
~~  $\alpha = 50$ & \textit{He He "It The ... You It "[They They) THE...I I HE IT [There There THEYRE WE "A A FOR "[...]} & \MT & 0.14 \\
\contpt \\~~ $\alpha = 5$ & \textit{The stable was then shut, leaving Pinocchio all alone. He had not eaten for many hours and began to yawn from great hunger. As he yawned, he opened his mouth wide, like an oven.} & \Htwo & 0.84 \\

\bottomrule[1.5pt]
\end{tabular}}
    \caption{Example paragraphs from ``Pinocchio'' (IT $\rightarrow$ EN) translated with \Htwo personalization. Setup outputs are from Gemma 2 9B; \expt, MS, \contht and \contpt (both $\alpha = 5$) can flip the \classifier{} prediction.}
    \label{tab:examples-all-it}
\end{table*}

\begin{table*}
    \small
    \centering
    \scalebox{0.9}{
\begin{tabular}{p{0.12\textwidth}p{0.70\textwidth}C{0.1\textwidth}C{0.08\textwidth}}
\toprule[1.5pt] 
Source (FR) & Cette somme vous sera restitu\'ee \'a votre sortie de prison, dit le juge. En attendant, vous êtes libres sous caution. & \classifier{} & \comet{} \\
\midrule
\textcolor{pastelorange}{\Hone} & \textit{This sum of money will be returned to you when you leave prison,' said the judge. 'In the meantime you are free on bail.} & - & - \\[4pt]
\textcolor{pastelorange}{ZS} & \textit{This amount will be returned to you upon your release from prison, the judge said. Meanwhile, you are free on bail.} & \MT & 0.79 \\[4pt]
\midrule
\contht \\~~ $\alpha = $ 5 & \textit{This sum will be repaid to you at your departure from prison, the judge declared. In the meantime, you are released under bond.} & \MT & 0.85 \\
~~ $\alpha = $ \textcolor{purple}{50} & \textit{\#echo all ought to goodness too was put moreover goodness behaving goodness goodness goodness goodness } [...] & \Hone & 0.20 \\
~~  $\alpha = $ \textcolor{purple}{150} & \textit{ GURL callers goodness\_Noneummy amid\'ala Reportedaps Beaverishlor Decompiled\_unset queer headquarters Headquarters twe ... pocket-or callers dt behaving indoorsgetRepository callers\_lazy uninsured tipostiFRING callers callers callers callers Guinness\_None ... foss bedroomsitud facet goodness goodness goodness} [...] & \Hone & 0.23 \\

\midrule
\toprule

Source (DE) & Nein – was Sie lieber haben, Herr Leutnant! Nur keine Zeremonien, es ist doch ganz einerlei. & \classifier{} & \comet{} \\
\midrule
\textcolor{pastelorange}{\Htwo} & \textit{"No, no-whatever you would rather have, Lieutenant Hofmiller! Please don't stand on ceremony, it makes no difference to us."} & - & - \\[4pt]
\textcolor{pastelorange}{ZS} & \textit{No, whatever you prefer, Lieutenant! Just no ceremonies, it doesn't matter.} & \MT & 0.76 \\[4pt]
\midrule
\contht \\~~ $\alpha = $ 5 & \textit{No, anything at all you want, sir!  Just don't make a fuss about it, it really doesn't matter.} & \Htwo & 0.79 \\
~~  $\alpha = $ \textcolor{purple}{50} & \textit{">I Don't worry about that... I don't want a ceremony for this one. It's not important...} & \Htwo & 0.46 \\
~~  $\alpha = $ \textcolor{purple}{150} & \textit{IWhenInWhatItDonIf Sometimes AIs Celebrating cerimonies... Sosir please don't have parties ey'} [...] & \Htwo & 0.24 \\

\bottomrule[1.5pt]
\end{tabular}}
    \caption{Examples from different languages being classified as Human when using \textcolor{purple}{extreme} $\alpha$ values.}
    \label{tab:extreme-alphas}
\end{table*}

\end{CJK*}
\end{document}